\documentclass[runningheads]{llncs}
\usepackage[T1]{fontenc}
\usepackage{graphicx}
\usepackage{cite}
\usepackage{algorithmic}
\usepackage{textcomp}
\usepackage{xcolor}
\usepackage{subcaption}
\usepackage{booktabs}
\usepackage{float}
\begin{document}
\title{Machine Learning Data Suitability and Performance Testing Using Fault Injection Testing Framework}
\titlerunning{Machine Learning Performance Testing Framework}
%
\author{Manal Rahal\inst{1} \and
Bestoun S. Ahmed\inst{1,2} \and
Jörgen Samuelsson\inst{3}}
\authorrunning{M. Rahal et al.}
\institute{Dept of Mathematics and Computer Science, Karlstad University \email{manal.rahal@kau.se}\\
\and Department of Computer Science, Faculty of Electrical Engineering, Czech Technical University in Prague, Prague, 16627, Czech Republic\\
\email{bestoun@kau.se}
\and Dept of Engineering and Chemical Sciences, Karlstad University
\email{jorgen.samuelsson@kau.se}}

\maketitle              
\begin{abstract}
Creating resilient machine learning (ML) systems has become necessary to ensure production-ready ML systems that acquire user confidence seamlessly. The quality of the input data and the model highly influence the successful end-to-end testing in data-sensitive systems. However, the testing approaches of input data are not as systematic and are few compared to model testing. To address this gap, this paper presents the Fault Injection for Undesirable Learning in input Data (FIUL-Data) testing framework that tests the resilience of ML models to multiple intentionally-triggered data faults. Data mutators explore vulnerabilities of ML systems against the effects of different fault injections. The proposed framework is designed based on three main ideas: The mutators are not random; one data mutator is applied at an instance of time, and the selected ML models are optimized beforehand. This paper evaluates the FIUL-Data framework using data from analytical chemistry, comprising retention time measurements of anti-sense oligonucleotide. Empirical evaluation is carried out in a two-step process in which the responses of selected ML models to data mutation are analyzed individually and then compared with each other. The results show that the FIUL-Data framework allows the evaluation of the resilience of ML models. In most experiments cases, ML models show higher resilience at larger training datasets, where gradient boost performed better than support vector regression in smaller training sets. Overall, the mean squared error metric is useful in evaluating the resilience of models due to its higher sensitivity to data mutation. 

\end{abstract}
\section{Introduction}
The world is experiencing rapid evolution in using artificial intelligence and machine learning (ML) in almost every domain. This trend has raised questions about the resilience of ML systems to data faults, most importantly in safety-critical applications such as autonomous driving, cyber security, and healthcare \cite{ZhangJie2022}. In light of these concerns and the serious consequences of failure in ML systems, testing methods have gained much attention in the research community and industry \cite{Ma2018}. Furthermore, the resilience of ML systems has become an important requirement to gain users' trust \cite{Jha2018}. Given its unpredictable behavior, testing ML systems is more complex than classical software since inspecting the ML algorithm alone is not sufficient \cite{Breck2017, Nurminen2019}. The behavior of the ML system does not depend solely on the algorithm but also on the training and testing data, the choice of hyperparameters, and the optimizer. All these factors impact the performance of the system \cite{Riccio2020}. The influence of data on the performance of the ML system is not negotiable. Therefore, performing tests to evaluate the suitability of the data is equally critical to achieving a production-ready ML system.

In the literature, various systematic testing methods are effective in testing the ML model, such as mutation testing (MT) and black-box testing tools \cite{Narayanan2021}. However, not as many systematic methods were investigated to evaluate the training data as stated by Narayanan \emph{et al.} \cite{Narayanan2021}. Having an important impact on the performance of the ML, major faults in the input data could lead to incorrect outcomes by the system. In some cases, not faults, but drifts in the real-time data lead to undesirable outcomes. Fault injection is one of these few available methods that aim to intentionally inject faults into the data, as described by \cite{Riccio2020} as data sensitive faults, in an attempt to change the behavior of the system \cite{Gangolli2022}. The more an ML system is resilient to data-sensitive faults, the better a system can learn from incomplete data and unexpected observations. Therefore, models that have high resilience generally generalize better to far-from-perfect real-world data.

ML systems based on supervised learning algorithms often assume the input data is static. However, this assumption does not necessarily hold when the system is deployed in the real world \cite{Katzir2018}. Therefore, unexpected events are likely to occur and can cause risks to the performance of the ML system. But a resilient ML system, as described by \cite{Vairo2023}, has the capacity to absorb data fluctuations without performance degradation. According to \cite{Vairo2023}, a resilient system can monitor, learn, anticipate, and respond to adversity. This means that the system should be able to maintain good performance when the input data are disrupted, to a certain extent. The sources of data disruption are many and could be classified as natural, system-related, or human errors, and can also include external factors \cite{Vairo2023}. The diversity in the types of faults raises questions such as are some ML models more resilient to data faults than others and which faults have the biggest influence?

Like any ML system, multiple fault sources can influence the data collected from a chromatography system. Consistent records of data could be attributed to the performance of the chromatographic instruments, experimental conditions, and other external factors. Such variations and errors are common in any real-world application and lead to degradation in the performance of the ML system. Therefore, there is a need for ML models that can predict under complicated uncertainty, yet perform efficiently when used in decision-making \cite{Lotfi2022}. To address this gap, we present the Fault Injection for Undesirable Learning in input Data (FIUL-Data) testing framework to evaluate the resilience of an ML system to common faults. The design intends to introduce likely-occurring faults through artificial mutators before applying the ML model. The proposed framework is generalizable, explainable, feasible on a scale, and applicable in multiple domains. The main ideas behind designing the FIUL-data framework are (1) the mutators are not random; they are formulated based on previous knowledge about the data, (2) it is a single fault application; so that at any instance of time only one fault is applied, and (3) the ML models used in the evaluation phase are selected based on their suitability to the dataset. FIUL-Data empirically validates the resilience of ML systems by introducing data faults to the ML input data in scenarios that might otherwise be rarely considered. FIUL-Data framework is evaluated on a case study from the analytical chemistry domain. The data set includes antisense oligonucleotide sequences (ASOs) and their experimentally observed retention time ($t_\mathrm{R}$). These text-coded sequences are transformed into numeric features before applying any ML system \cite{OREILLY2018}. The data set is collected through a sequence of chromatography experiments from the Chemistry Department of Karlstad University. The evaluation of FIUL-Data consists of two steps; first, each ML model is evaluated individually. In the second step, the models are compared with each other to have a comparative view of the vulnerabilities of each model against the different fault injections. We are mainly considering the supervised ML model in our case study. However, the proposed framework could apply to other types of ML systems.

In this paper, we propose and evaluate the FIUL-Data framework usable in multiple domains against likely occurring data faults. As a result, this paper makes the following contributions:

\begin{itemize}
    \item Propose FIUL-Data as a data-mutation-based framework integrated with ML to evaluate the resilience of ML systems to data faults.
    \item Design and implement two data-level mutators applied to introduce likely-occurring faults to ML input data.
    \item The proposed FIUL-Data framework is evaluated on a dataset from the analytical chemistry field to demonstrate the usefulness of the framework in a real-world application.
    \item Propose and perform a multi-metric evaluation of the FIUL-Data framework to enable quantitative evaluation of the metrics and generation of insights.
\end{itemize}

The remainder of this paper is structured as follows: Section \ref{sec:background} summarizes the relevant background concepts and lays out the necessary terminologies to understand the paper. Section \ref{sec:methodology} details the research questions and the experimental setup to apply and evaluate the proposed framework. The evaluation results from the use case and the answers to the research questions are presented in Section \ref{sec:results}. Finally, Section \ref{sec:conclusion} concludes the paper with a summary of the findings.

\section{Background}\label{sec:background}
This section provides a selected overview of relevant MT definitions and applications in the literature.

\subsection{ML in Chromatography Applications}
Chromatography is an important separation method that is used in all chemistry fields \cite{Fornstedt2015}. Chromatography is considered powerful in separating mixtures of compounds even with similar physical properties due to the large number of partitioning steps involved \cite{Fornstedt2015}. The output of the separation is gaussian-shaped peaks for each eluted compound in the mixture. ML is commonly used in chromatographic separation applications to predict experiments before they are conducted in the laboratory. This is possible by predicting important parameters, such as $t_\mathrm{R}$, which is the time a compound spends in the system from injection to elution. One of the most important goals in chromatographic separation is to achieve a sufficiently high resolution between the eluted peaks within a reasonable experimental time and resources \cite{Korany2012}. Therefore, optimizing the experiment conditions to achieve this goal requires much effort. However, the optimization task can be complicated and time-consuming, given the large space of experimental variables and the possibilities of interaction among them \cite{Webb2009}. The accurate performance of ML predictive models allows analytical chemists to reduce the costly and time-consuming experiments needed to achieve optimal separation conditions. In such cases, peak resolutions or $t_\mathrm{R}$ could be predicted as the output. Once the space of chemical conditions is controlled, more efficient separation experiments can be conducted in the laboratory. In the literature, various ML models have been tested in this context, including artificial neural networks (ANN) and traditional models such as in \cite{ENMARK2022, Kohlbacher2006, Strum2007}, which have shown promising results.

Another important application of ML in chromatography is to use it to analyze the chromatographic data that result from the experiments. Regardless of the type of chromatography system, data analysis is time-consuming and often requires manual human intervention \cite{Risum2019}. Therefore, researchers actively investigate the potential of ML to reduce human-dependent steps in the analysis stage using mainly ANN, as in \cite{Tran2007, Risum2019, Archivio2019, Petritis2003}. For example, \cite{Risum2019} tested an approach based on convolutional neural networks to automatically evaluate the modeled elution profiles of the gas-chromatographic data. In \cite{Archivio2019}, a multilayer neural network (NN) was applied to predict the retention behavior of amino acids in reversed-phase liquid chromatography. The authors concluded that NNs are powerful in modeling the influence of various gradient elution modes. ML offers great potential to advance the separation tasks toward more efficient handling of chromatographic data from collection to analysis. In this paper, we provide another example of the usefulness of ML with a focus on building more resilient separation pipelines in chromatography.

\subsection{Common Faults in Chromatography Data}
The reliability of the data generated by chromatography experiments has been investigated greatly in the literature. Although modern instruments used in laboratories have user-friendly interfaces, some types of variation and errors are inevitable \cite{Kuselman2013QA}. Multiple approaches have been proposed to classify and quantify the errors in an attempt to understand their influence on the certainty of the measurements. In general, variations can be seen as person or system caused. Guiochon \emph{et al.} \cite{Guiochon1988} discuss the different sources of errors and their influence on chromatographic measurements. Kuselman \emph{et al.} \cite{Kuselman2013} identify nine human-related errors while performing experiments. The most common sources of common systematic errors in chromatography were also investigated in \cite{Kaiser1971}. The causes of systematic errors ranged from errors in sample handling to wrong evaluation and interpretation of results. In the case of ASO chromatography experiments, errors are reflected in the output data in the form of skewness in the detected peaks, low signal-to-noise ratio, and high variation of the outcome across replicates of the sample. The inevitability and recurrence of errors in chromatography, demand handling before data are used for analysis or ML purposes \cite{Hellier2001}. MT methods integrated with ML are one of the approaches used to simulate likely-to-happen errors and study their influence on the performance of the models.

\subsection{Data Mutation}\label{sec:datamutation}
To ensure the systems work as intended in real but uncertain conditions, we must understand and consider the faults in our input data \cite{Nurminen2019}. There is no doubt that different types of faults would have a different influence on the performance of the ML system. Many approaches deal with the different faults, such as changing the input data, changing the model, or building a resilient system to faults \cite{Nurminen2019}. One of the approaches to understand the influence of certain errors is the application of designed artificial mutations to the input data. 

Originally, MT is a popular software engineering (SE) technique widely used in academia and industry \cite{papadakis2019}. The concept of MT is based on using artificial faults for system testing purposes. In other words, MT for evaluation is used to measure the effectiveness of a system in finding faults \cite{Zhu2018}. Traditionally, MT application involved inserting individual faults into the software, but later approaches tested higher-order mutations in which multiple faults are injected at once, which was shown to be expensive \cite{Ghiduk2017}. Although MT is a proven technique in SE, it recently started gaining attention in the field of ML, specifically in the subfield of deep learning (DL) \cite{Tambon2023}. However, MT for ML systems is still considered to be in the early stages \cite{LU2022}. The interest in adopting established methods such as MT is triggered by the willingness to improve the trustworthiness of DL systems \cite{Tambon2023}. It should be noted that by concept, MT has been most commonly used to test models, but rarely at the data level, which is our paper's focus. 

Motivated by the success of MT in classical software systems and recently in DL models, the FIUL-Data framework applies mutation at the data level. In the literature, the tools used to evaluate the input data for ML are limited \cite{Narayanan2021}. To address this gap, we propose the FIUL-Data framework to assess the responsiveness of multiple ML models to input data mutations. The framework is applied to ASO chromatographic data where the designed data mutations represent common faults in the field.

\section{Methodology}\label{sec:methodology}
In this section, we describe the systematic approach followed to study the resilience of ML models to data faults, including the specific research questions that guided the experiments. The dataset used to evaluate the FIUL-Data framework and the methods used for the evaluation are also described in detail.

\subsection{Research Questions}\label{sec:RQs}
This paper aims to answer the following research questions (RQs):

\begin{itemize}
    \item \textbf{RQ1: How does the reduction in training data influence the resilience of the ML model?}
    \item \textbf{RQ2: How does the selection of the size of a certain class of data influence the resilience of the ML model?}
    \item \textbf{RQ3: How to evaluate the resilience of different ML models in response to different data mutators?}
\end{itemize}

\subsection{Use Case Dataset}\label{sec:dataset}
The data used in our use case are obtained by means of a chromatography experiment aimed at separating impurities from an ASO compound. The experiments were carried out under two different chemical conditions, therefore resulting in two different datasets G1 and G3. During the separation process, an aqueous-organic mixture is continuously pumped into the chromatography column, where the amount of organic solvent in the mixture increases over time (gradient time); this is called the gradient. The first dataset (G1) is collected from experiments in which the gradient equals 11 minutes. The second dataset (G3) is collected at a gradient of 44 minutes. The change in gradient results in a new $t_\mathrm{R}$ for each unique compound entering the chromatography system. At higher gradients, ASO compounds are retained longer in the system. Therefore, data collected at higher gradients are considered more sensitive to slight changes in experimental conditions. 

In chemistry, the ASO compound is represented in a combination of four different nucleotide bases, adenine (A), thymine (T), cytosine (C), and guanine (G), forming a sequence. In this case study, non-phosphorothioated known as native and phosphorothioated ASO sequences are collected in a dataset with their respective $t_\mathrm{R}$ as the target variable. The $t_\mathrm{R}$ is always recorded as the separated compounds individually exit the system. 

\begin{figure}
\centering
    \includegraphics[scale=.35]{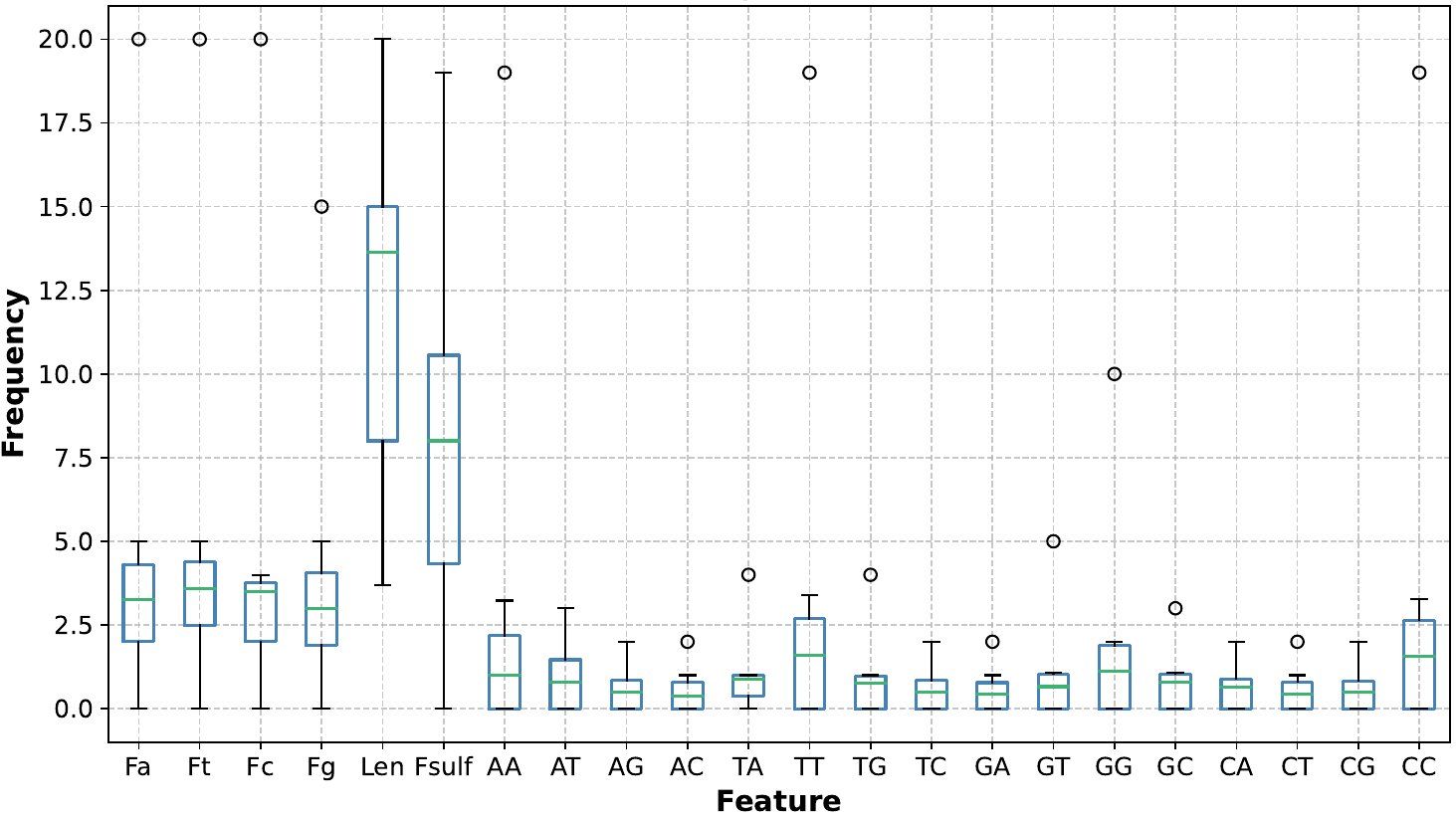}
    \caption{Frequency of features in G3 training data. The features representing unordered di-nucleotides were removed for visualization purposes.}
    \label{fig:G44trainboxplot}
\end{figure}

After pre-processing the datasets and removing the incomplete records, the clean G1 dataset included 876 data points and G3 dataset had 870 data points. Both datasets have compounds that do not include sulfur (non-phosphorothioated), partially phosphorothioated compounds, and others fully phosphorothioated. Both datasets have more than 79\% of the compounds partially or fully phosphorothioated. As part of the data preparation methods before applying ML, the nucleotide sequences are encoded into numeric values, referred to as features, such as the frequency of each nucleotide and di-nucleotide (ordered and unordered) in a sequence, the total length, and the number of sulfur atoms present. Figure \ref{fig:G44trainboxplot} shows the frequency range for the encoded features in the G3 training dataset. The same encoding system is applied to the testing data. As seen in Figure \ref{fig:G44trainboxplot}, the ASO sequence can reach 20 nucleotides long, whereas, the number of sulfur atoms in the sequences varies between 0 and 19 atoms. The occurrence of A, C, T, and G nucleotide bases is relatively similar, while, in di-nucleotide occurrence, TT and CC are the most frequent.

\subsection{The FIUL-Data Framework}\label{sec:experimentsetup}
\begin{figure}
    \centering
    \includegraphics[scale=.35]{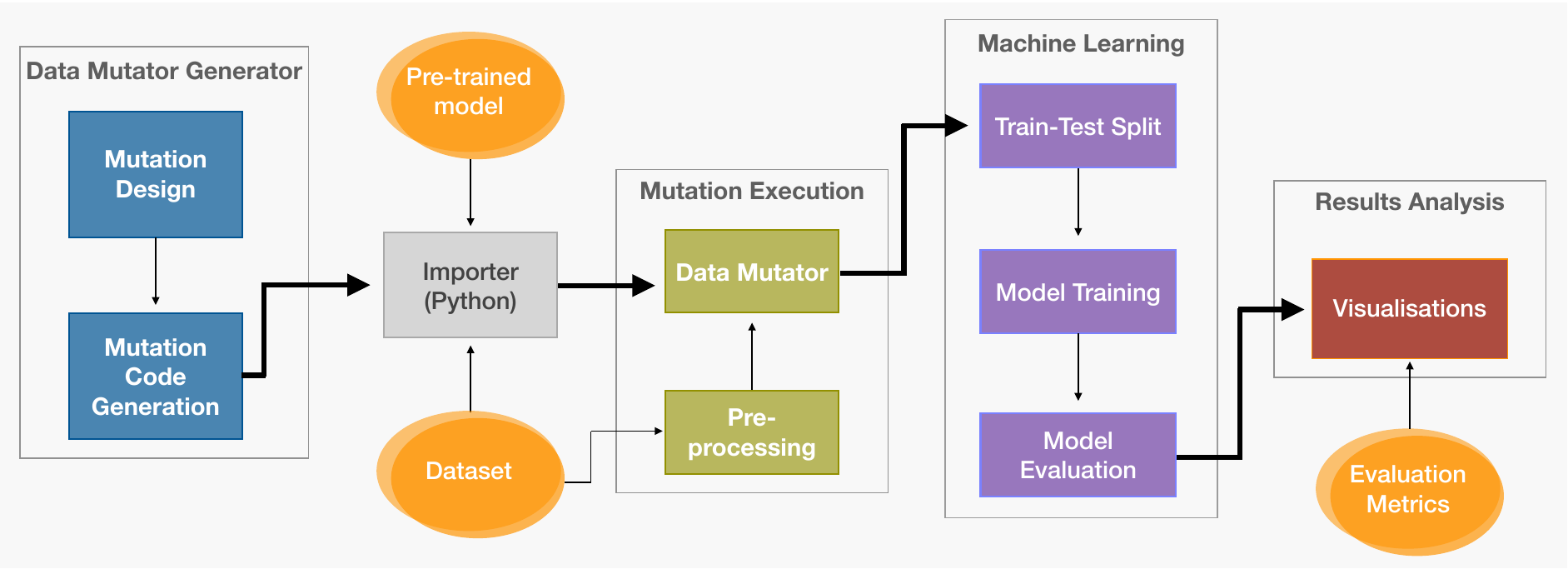}
    \caption{Conceptual illustration of the FIUL-Data framework}
    \label{fig:framework}
\end{figure}

The FIUL-Data framework is built on four key phases, as illustrated in Figure \ref{fig:framework}. In the first phase, the data mutators are designed and coded. The operation of the data mutators is application-specific and relies on the common faults encountered during implementation. The execution of data mutators comes next, where pre-trained ML models and the programmed data mutators are imported and performed on clean data. Once the mutated data are ready, the ML cycle starts, including typical training and evaluation of the models based on consciously selected metrics. The results are integrated into useful visualizations in the last phase, and insights are concluded. In our case study, Python language is selected to implement the data mutators' functions and run the experiments.

The application of FIUL-Data framework consists of two general-purpose data mutators, the reduce\_data\_mutator and the select\_data\_mutator. The operation of both data mutators is described in detail in the following section. We note that the same data mutators are expected to behave differently in different applications. In this application, 10\% of the data is repeatedly removed in every iteration until only 10\% of the original data is left. The selected step-size induces small changes in the data that allows to record results and visualize behavioral trends from 10 data mutation iterations. In other words, the 10\% ieterative variation ensures that we have a sufficient number of instances from small yet sensitive changes that allows to monitor the behavior of the models. We note that the records in the dataset are randomly shuffled before applying the FIUL-Data framework. This is an important step, as compounds and impurities could share similar characteristics in the case of chromatography data. Data shuffling ensures representative distribution of the different compounds in the partitions of the training and testing sets. 

\begin{figure*}
\centering
    \includegraphics[scale=.3]{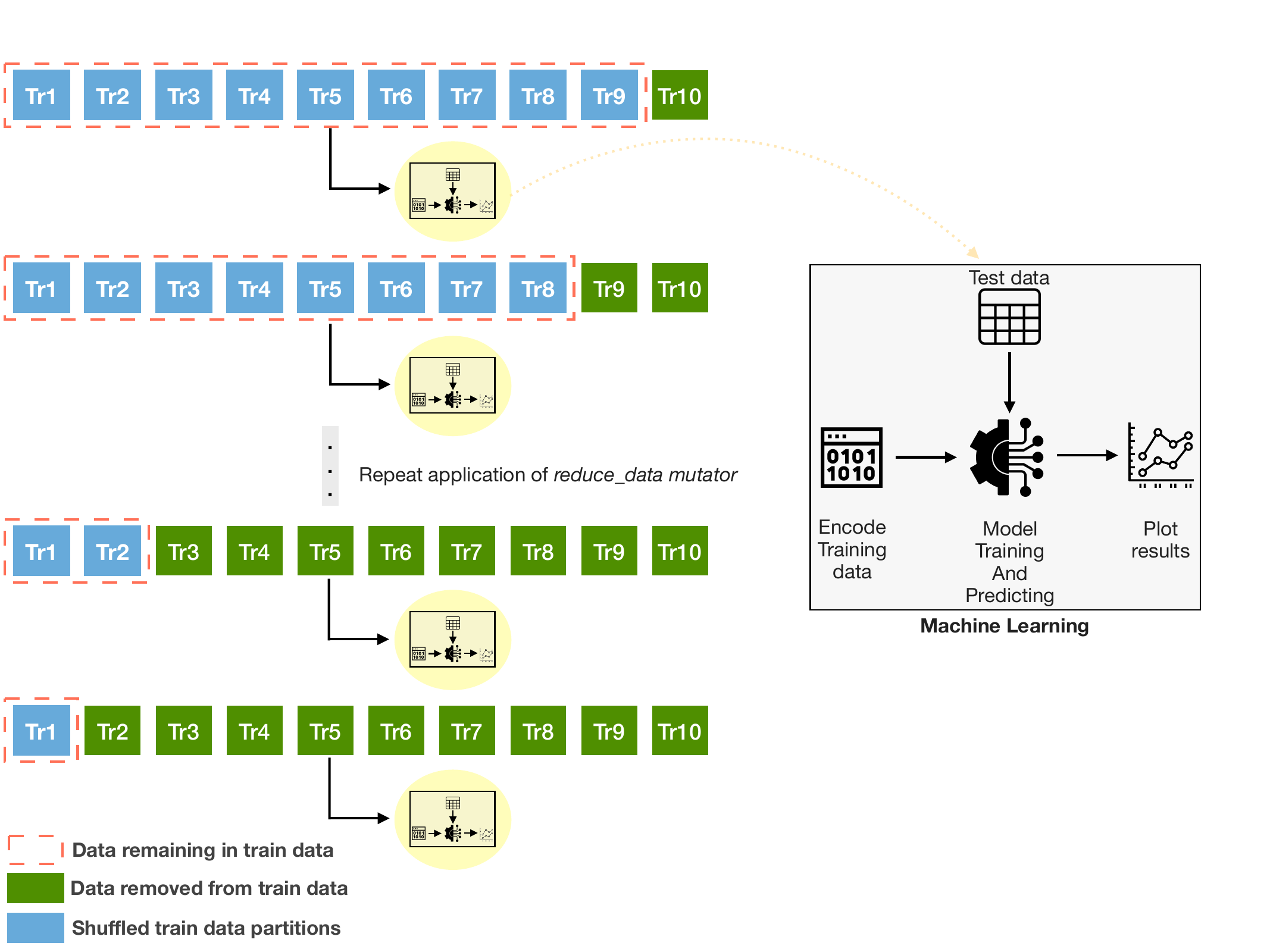}
    \caption{The operation of reduce\_data\_mutator }
    \label{fig:reducemutator}
\end{figure*}

\begin{figure*}
    \centering
    \includegraphics[scale=.3]{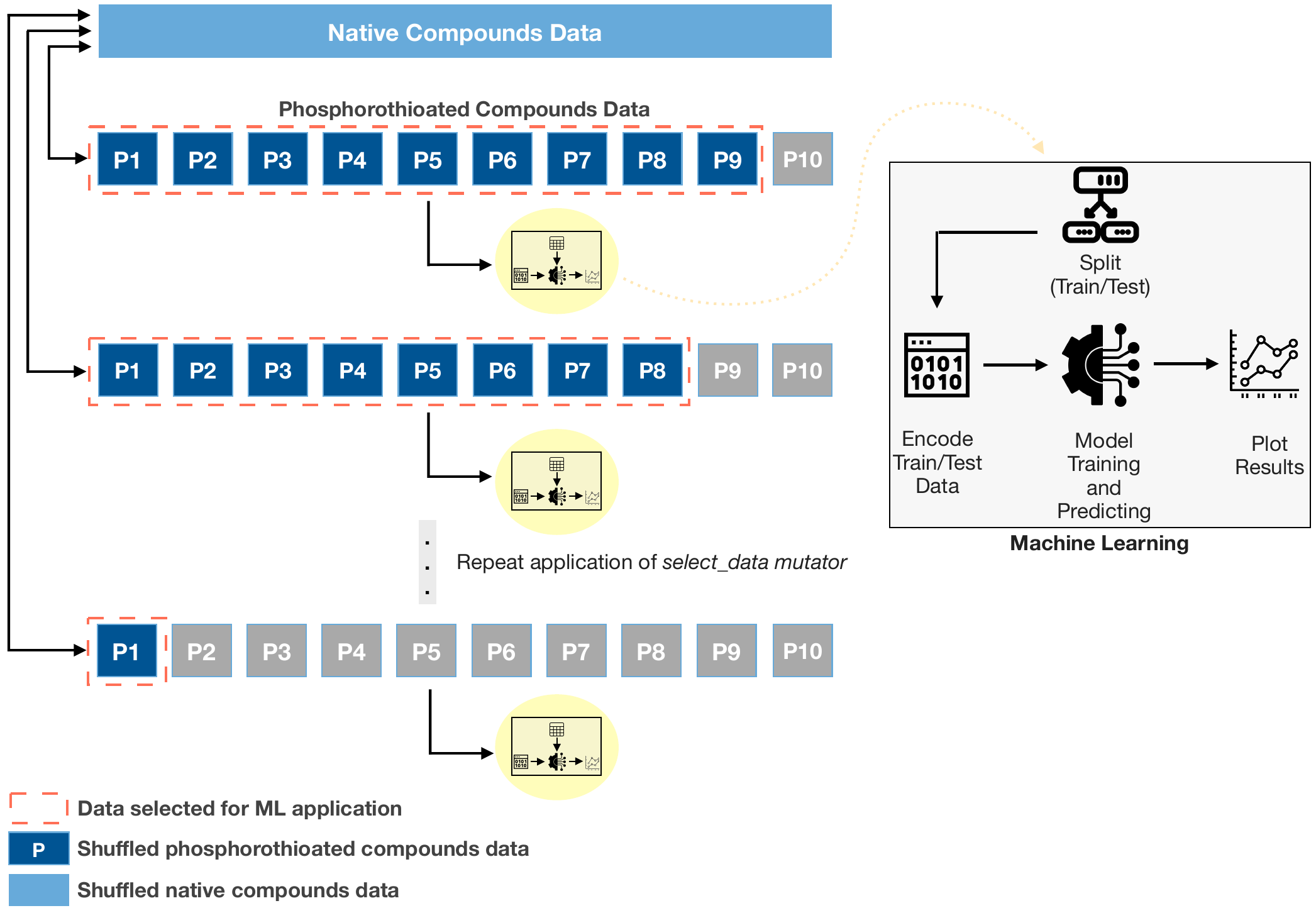}
    \caption{The operation of select\_data\_mutator }
    \label{fig:selectmutator}
\end{figure*}

\begin{itemize}
    \item \textbf{Reduce\_data\_mutator} The reduce\_data\_mutator reduces the training data by 10\% in each iteration. In some applications, where collecting data for ML is expensive and time-consuming, it is critical to know the size of training data that is sufficient for a good-performing model. Large data are a relative term and depend on the application being studied. In the case of chromatography data, laboratory experiments are expensive, the products used to perform the experiments are costly, and the experiments run for a long time. Therefore, collecting sufficient data could significantly reduce time and cost burdens. In our experiment, the data is first split into train and test datasets, 80\% and 20\%, respectively. The testing data remains unchanged, while the training data is reduced iteratively by 10\%. For every iteration, the model is fitted to the training data and $t_\mathrm{R}$ is predicted on the unseen data. At the end of each iteration, the coefficient of determination (R\textsuperscript{2}) train, R\textsuperscript{2} test, and the mean squared error (MSE) are recorded. The design and operation of reduce\_data\_mutator is illustrated in Figure \ref{fig:reducemutator}. 
        
    \item \textbf{Select\_data\_mutator} The select\_data\_mutator iteratively reduces a certain class in the data. In this case, the records of the compounds that lost one or more sulfur atoms from their sequence are removed at a rate of 10\% per iteration, while the other class (native compounds) remains unchanged. However, the percentage of the target class in the training and testing data after splitting is controlled for consistency purposes. The sulfur atom(s) loss is denoted by "-P=O" at the end of an ASO sequence. The models are then fitted to the new version of the training data and evaluated on the testing data. The operation of select\_data\_mutator is illustrated in Figure \ref{fig:selectmutator}. This mutator aims to reveal the influence of a certain class of data on the performance of the ML system. In this case study, the class of sequences having a -P=O suffix is subject to data mutation. 
\end{itemize}

\section{Results and Analysis}\label{sec:results}
After implementing the reduce\_data\_mutator and the select\_data\_mutator functions in a Python (3.9) supported framework, the G1 and G3 datasets along with the corresponding pre-trained and hypertuned ML models, were imported. Then, the FIUL-Data framework was applied to both datasets, where the performances of the Gradient boost(GB) and support vector regression (SVR) models in response to the type of fault injected were compared. The evaluation metrics used in monitoring the performance of individual models and when comparing models to each other were the MSE, the R\textsuperscript{2} train, and the R\textsuperscript{2} test. The MSE is chosen to observe the variation in the average squared difference between the predicted values and the observed values of $t_\mathrm{R}$. 

\subsection{Effect on G1 Dataset}
\begin{figure}
\centering
    \begin{subfigure}{0.4\textwidth}
         \includegraphics[width=\textwidth]{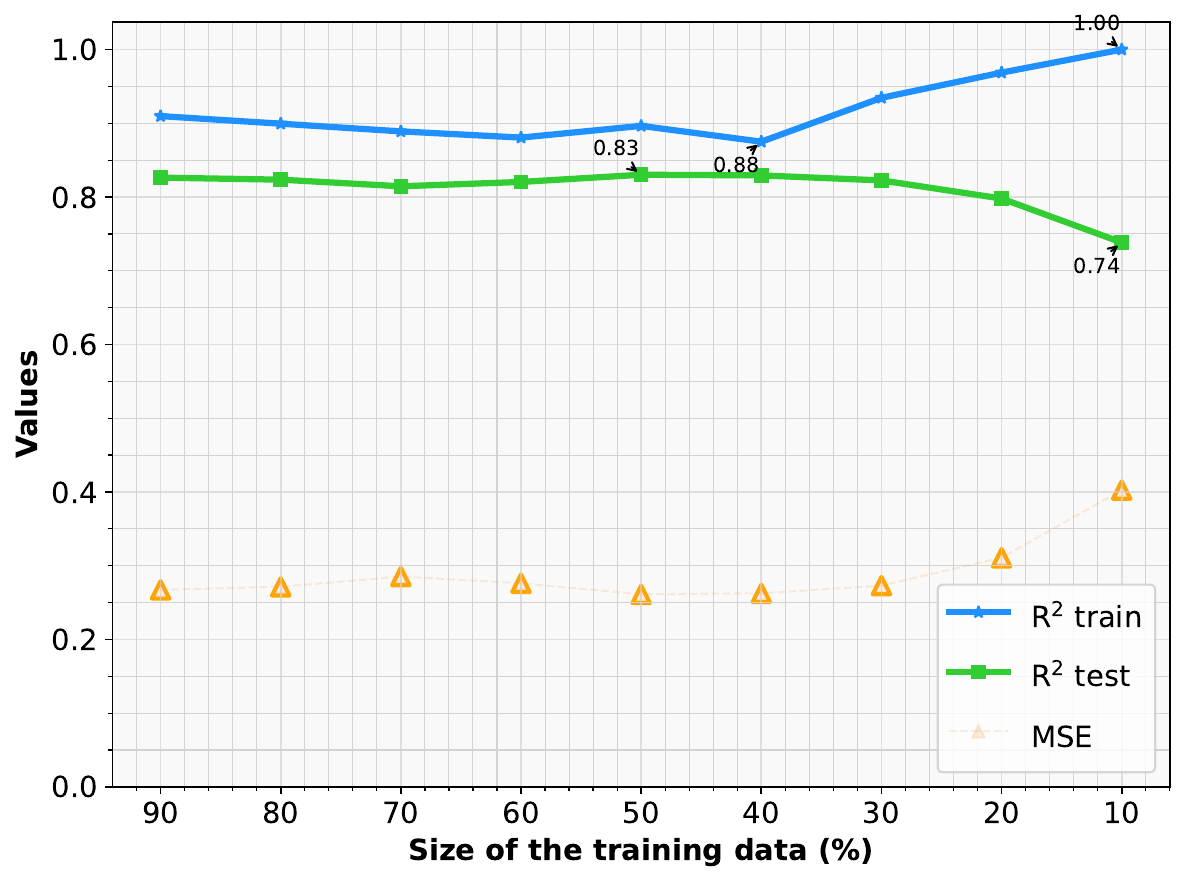}
         \caption{SVR}
     \end{subfigure}
    \begin{subfigure}{0.4\textwidth}
         \includegraphics[width=\textwidth]{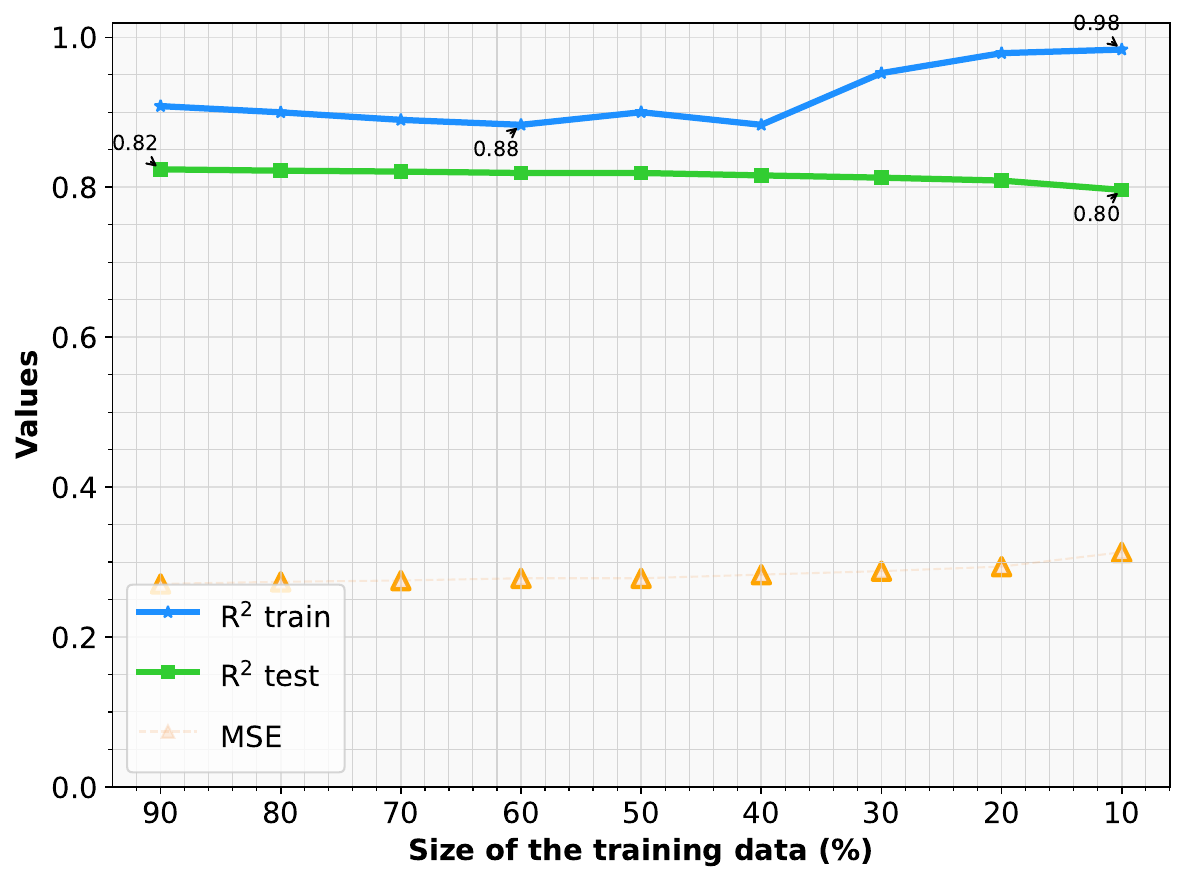}
         \caption{GB}
     \end{subfigure}
    \caption{The application of reduce\_data\_mutator in G1 dataset using SVR and GB models. Maximum and Minimum values are annotated.}
  \label{fig:GBSVR_G11}
\end{figure}

In Figure \ref{fig:GBSVR_G11}, the reduce\_data\_mutator is applied to the G1 dataset at a decreasing rate of 10\% in each iteration. For each of the mutation iterations, the accuracy of the train and the test are recorded in addition to the MSE values. The performance of the SVR and GB models on G1 data yields relatively good results. The accuracy on unseen data for the GB model ranges from 0.80 to 0.82, showing relatively stable performance against reducing the size of the training data. For the SVR model, the R\textsuperscript{2} test remained relatively stable until 60\% of the training data were removed, and the performance began to degrade, reaching a minimum R\textsuperscript{2} test of 0.74. Both models were shown to generalize reasonably well to unseen testing data when the training data is relatively large; however, GB performed better with a smaller training datasets. The same trend applies to MSE, which increased significantly for the SVR model with decreasing training data.

\begin{figure}
\centering
    \begin{subfigure}{0.4\textwidth}
         \includegraphics[width=\textwidth]{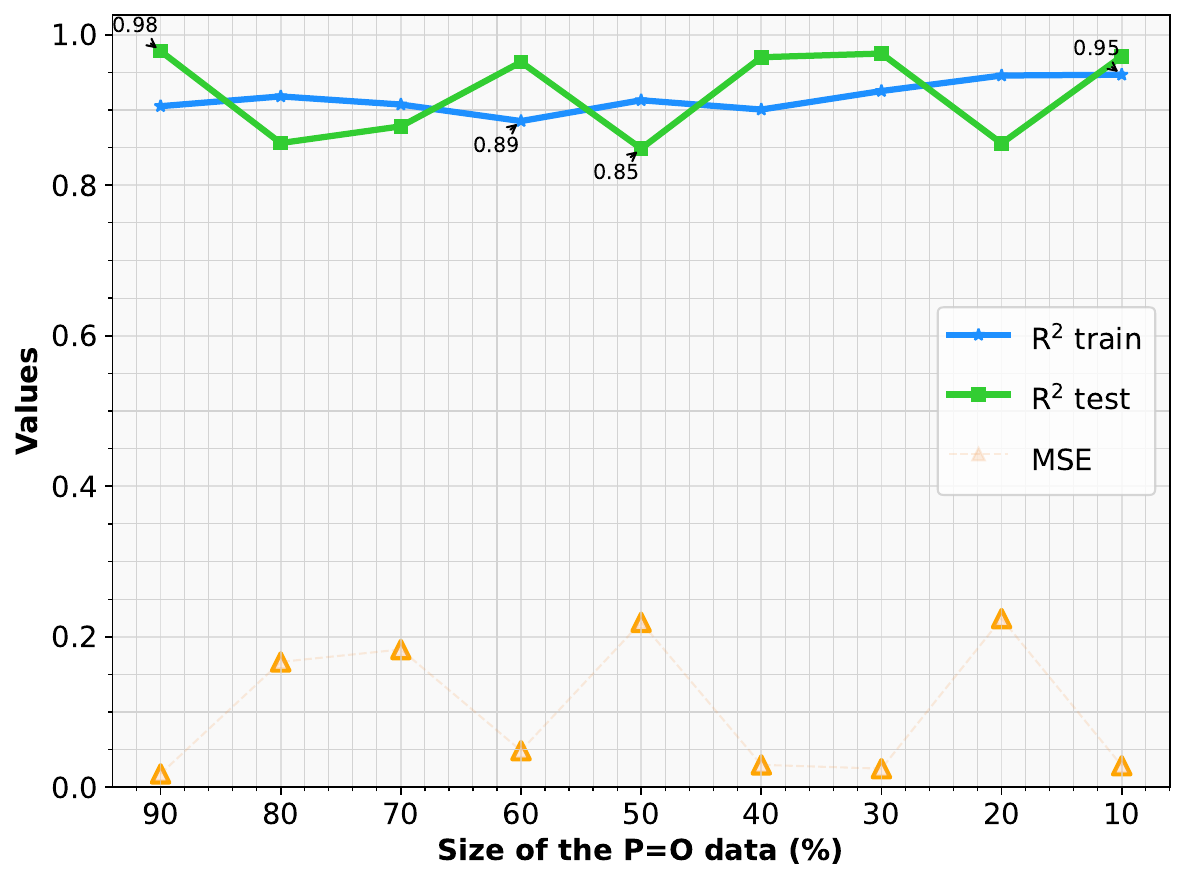}
         \caption{SVR}
    \end{subfigure}
    \begin{subfigure}{0.4\textwidth}
         \includegraphics[width=\textwidth]{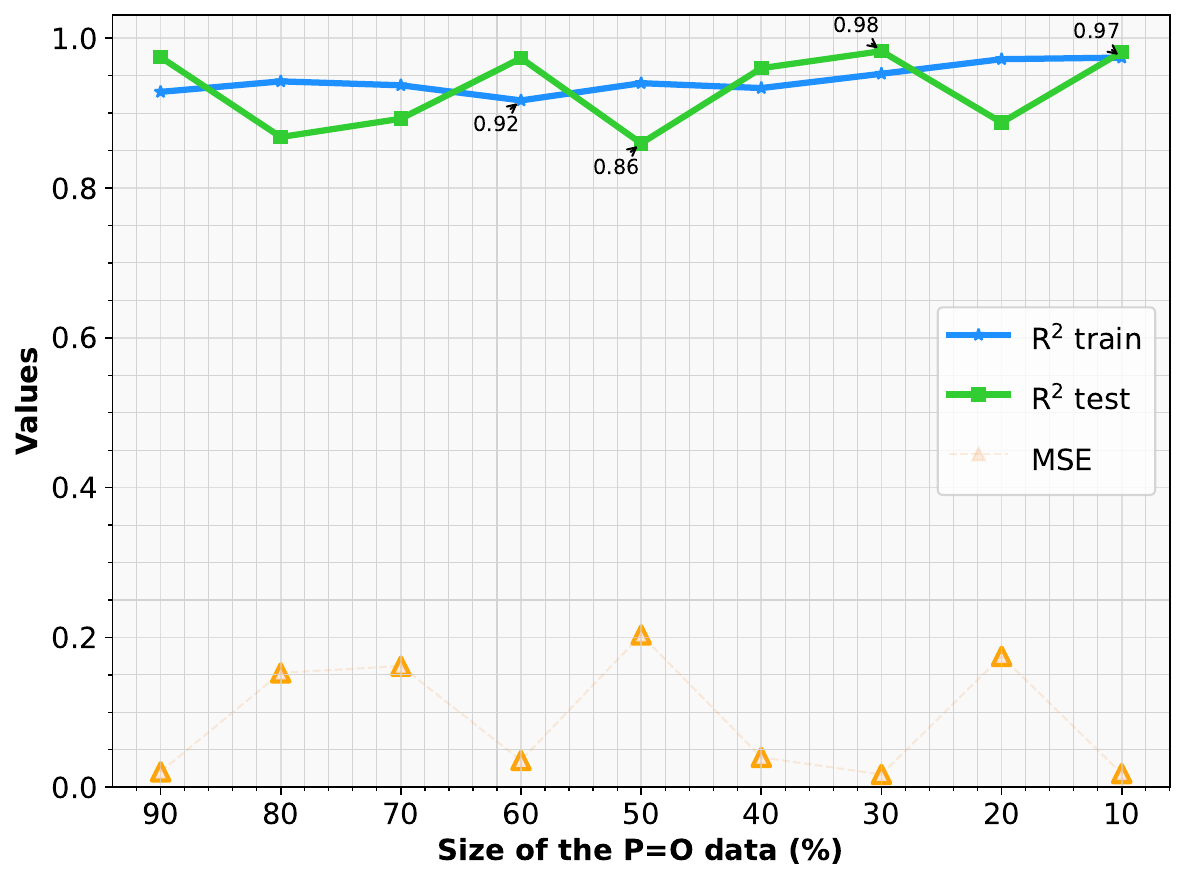}
         \caption{GB}
    \end{subfigure}
    \caption{The application of select\_data\_mutator in G1 dataset using SVR and GB models. Maximum and Minimum values are annotated.}
    \label{fig:GBSVR_G11_PO}
\end{figure}

The results of the application of select\_data\_mutator to input data are shown in Figure \ref{fig:GBSVR_G11_PO}. For the GB model, we can observe that the highest R\textsuperscript{2} test is achieved with a 30\% (0.98) subset size of the -P=O class. The R\textsuperscript{2} test ranges from 0.86 to 0.98 and the graph shows a trend of increasing R\textsuperscript{2} test with a decrease in the subset size of the -P=O data class. The GB and SVR models performed relatively similarly, with GB showing slightly better performance across iterations. However, high fluctuations are observed in the performance of both models in the case of select\_data\_mutator. The same trends are observed in the MSE values, ranging from 0.017 to 0.203 in GB and 0.018 to 0.029 in SVR. MSE is a measurement of error, so the lower the MSE value, the better the performance of the model. Both models achieved the lowest performance at the sizes of the subsets 90\% and 30\% of the -P=O class.

\subsection{Effect on G3 Dataset}
\begin{figure}
\centering
    \begin{subfigure}{0.4\textwidth}
         \includegraphics[width=\textwidth]{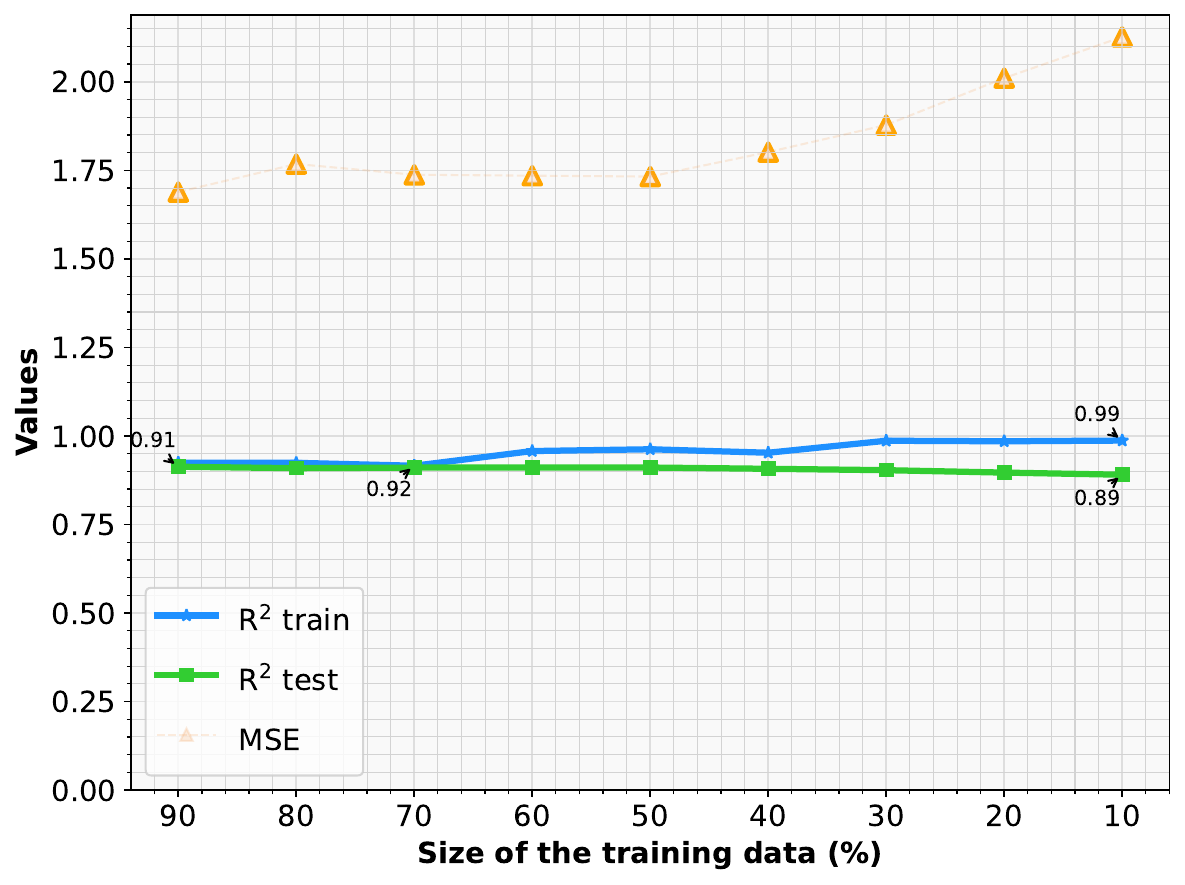}
         \caption{SVR}
    \end{subfigure}
    \begin{subfigure}{0.4\textwidth}
         \includegraphics[width=\textwidth]{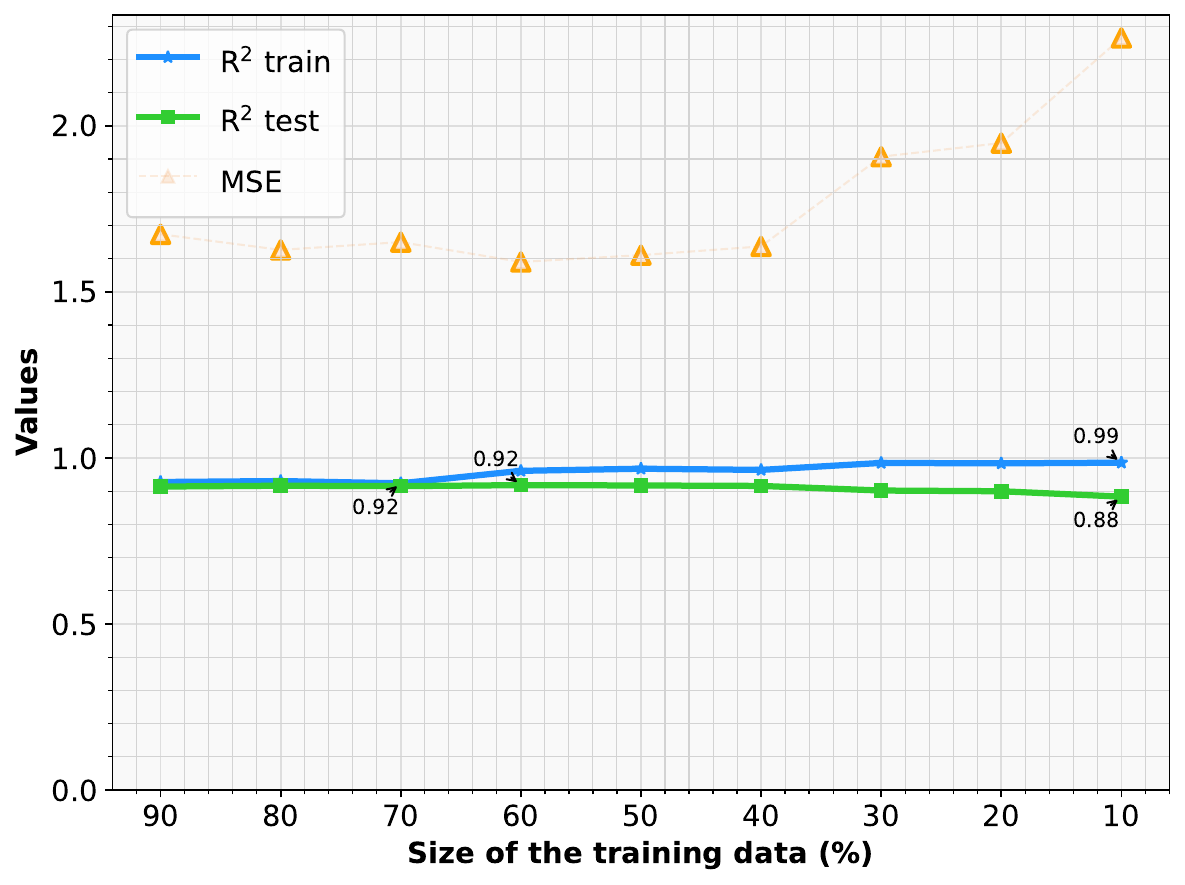}
         \caption{GB}
    \end{subfigure}
    \caption{The application of reduce\_data\_mutator in G3 dataset using SVR and GB models. Maximum and Minimum values are annotated.}\label{fig:GBSVR_G44}
\end{figure}

\begin{figure}
\centering
    \begin{subfigure}{0.4\textwidth}
         \includegraphics[width=\textwidth]{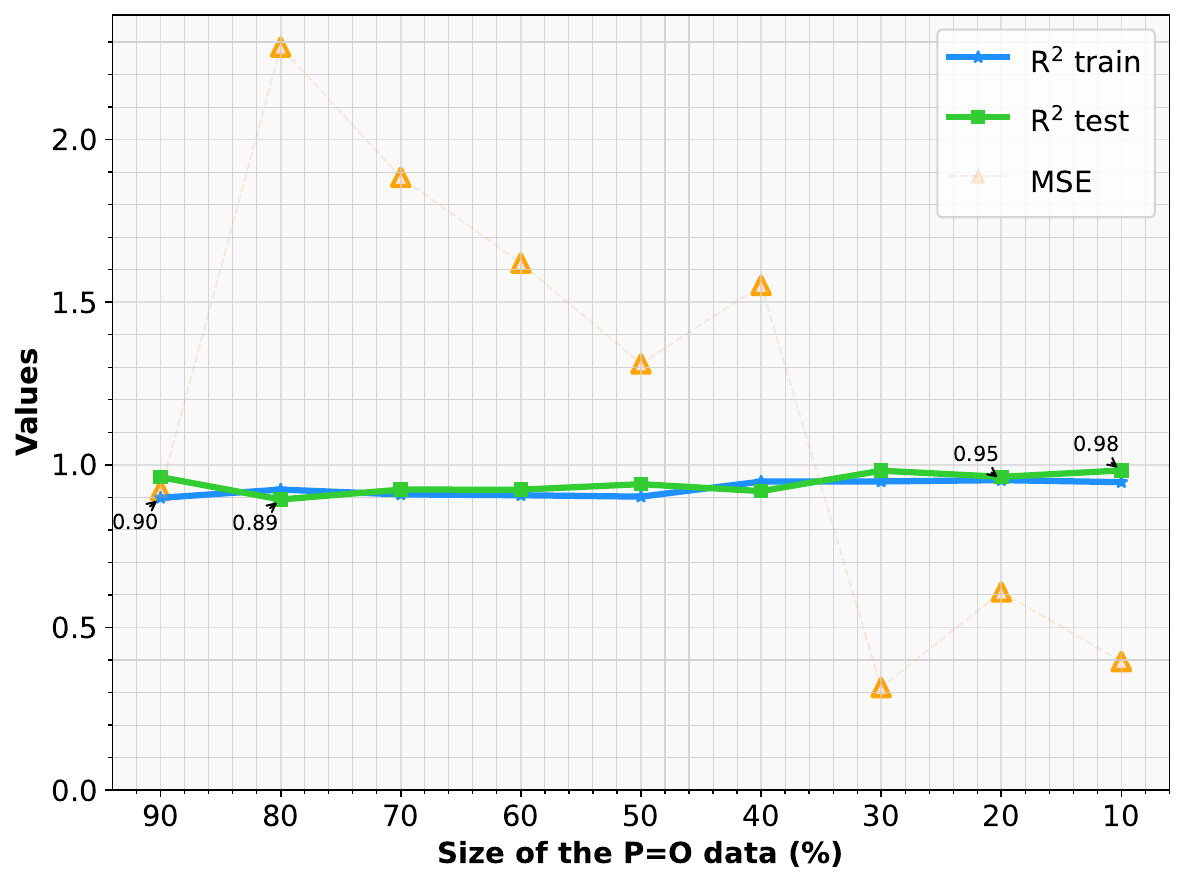}
         \caption{SVR}
    \end{subfigure}
    \begin{subfigure}{0.4\textwidth}
         \includegraphics[width=\textwidth]{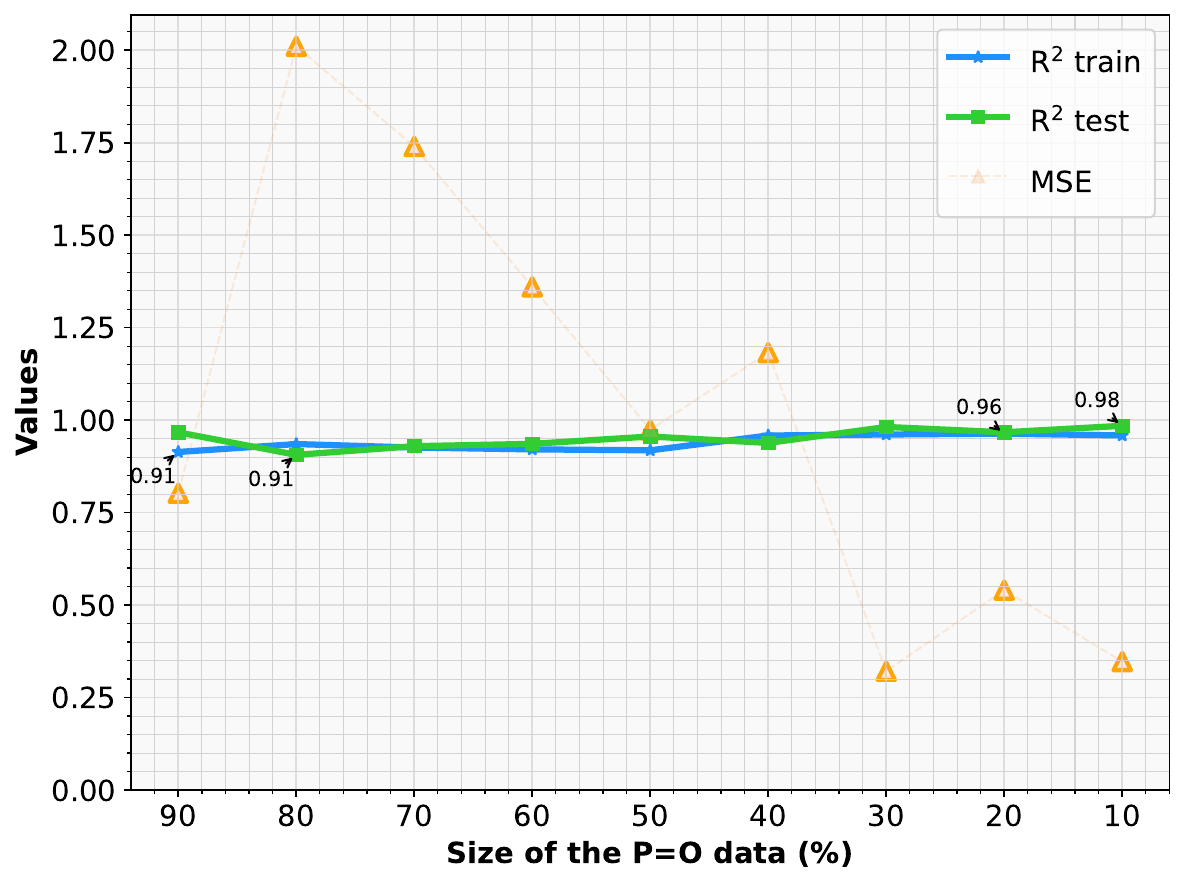}
         \caption{GB}
    \end{subfigure}
    \caption{The application of select\_data\_mutator in G3 dataset using SVR and GB models. Maximum and Minimum values are annotated.}
    \label{fig:GBSVR_G44_PO}
\end{figure}

Across both mutation applications and in almost all iterations, GB outperformed SVR as shown in the MSE trend line in Figures \ref{fig:GBSVR_G44} and \ref{fig:GBSVR_G44_PO}. Starting with the results of reduce\_data\_mutator, both GB and SVR models showed relatively high performance on unseen data, with the R\textsuperscript{2} test reaching a maximum value of 0.92 and 0.91 respectively. The R\textsuperscript{2} train for both models is consistently high, indicating effective learning during the training process. After 60\% of the training data is reduced, both models start to show over-fitting behavior where the training performance is exceptionally high, unlike the degrading R\textsuperscript{2} test values. The R\textsuperscript{2} test in both models remains stable until 70\% of the training data is removed, where the R\textsuperscript{2} test begins to show a downward trend. This trend is also reflected in the behavior of MSE, where it first shows a consistent trend, then a significant increase is observed after the 7\textsuperscript{th} iteration of the data mutation.

Figure \ref{fig:GBSVR_G44_PO} illustrates the results of the application of the select\_data\_mutator in G3 dataset. The values of R\textsuperscript{2} train and R\textsuperscript{2} test in the GB and SVR models are relatively similar and show stable behavior. However, a significant decrease in MSE is observed after the 2\textsuperscript{nd} iteration, reaching 1.31 and 0.9 in SVR and GB, respectively. Another steep decline is observed after the 6\textsuperscript{th} iteration where the error value decreases by 79.7\% and 73.7\% in SVR and GB, respectively.

\subsection{Effect Analysis}\label{sec:effectanalysis}
\begin{figure}
\centering
    \begin{subfigure}{0.4\textwidth}
         \includegraphics[width=\textwidth]{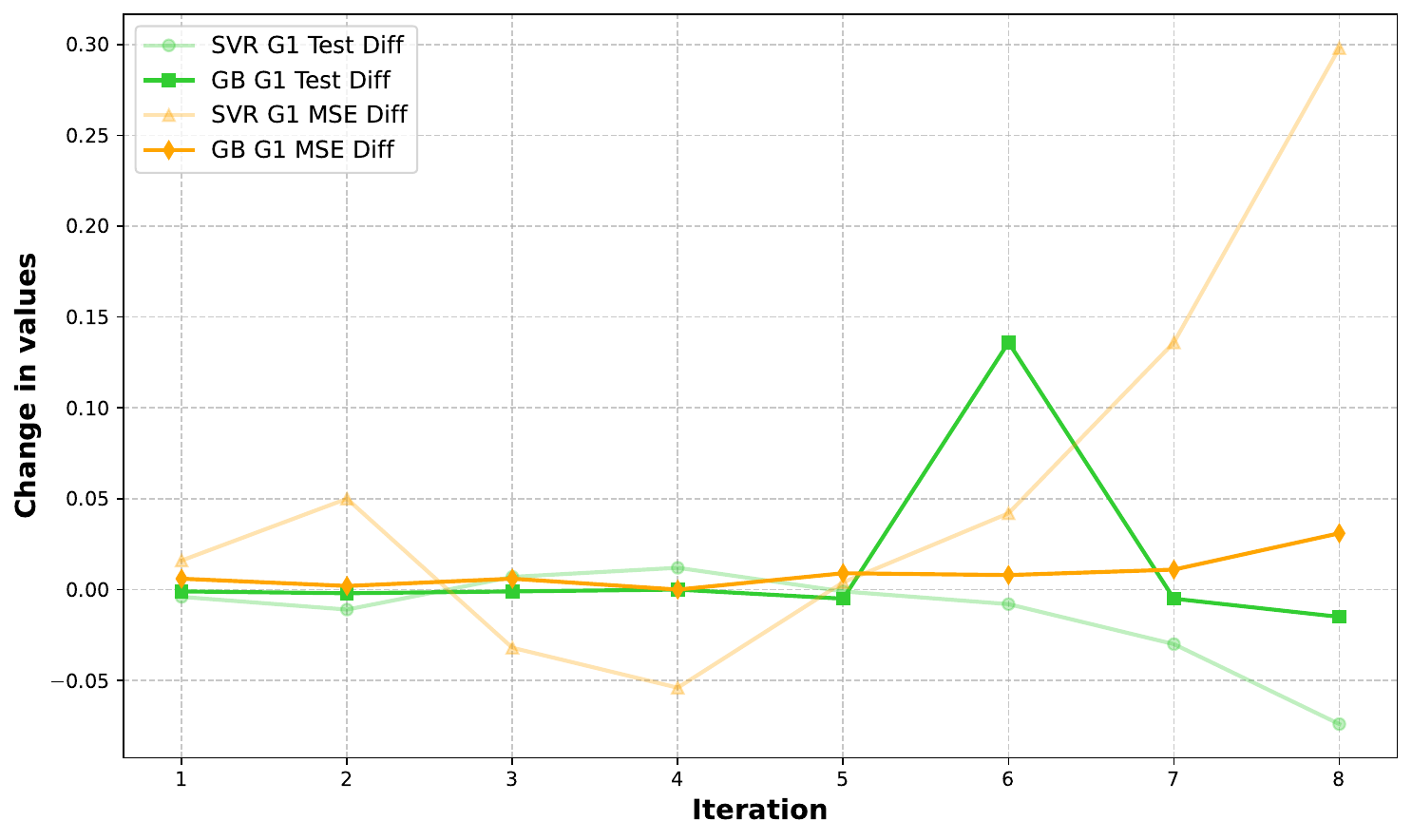}
         \caption{G1}
    \end{subfigure}
    \begin{subfigure}{0.4\textwidth}
         \includegraphics[width=\textwidth]{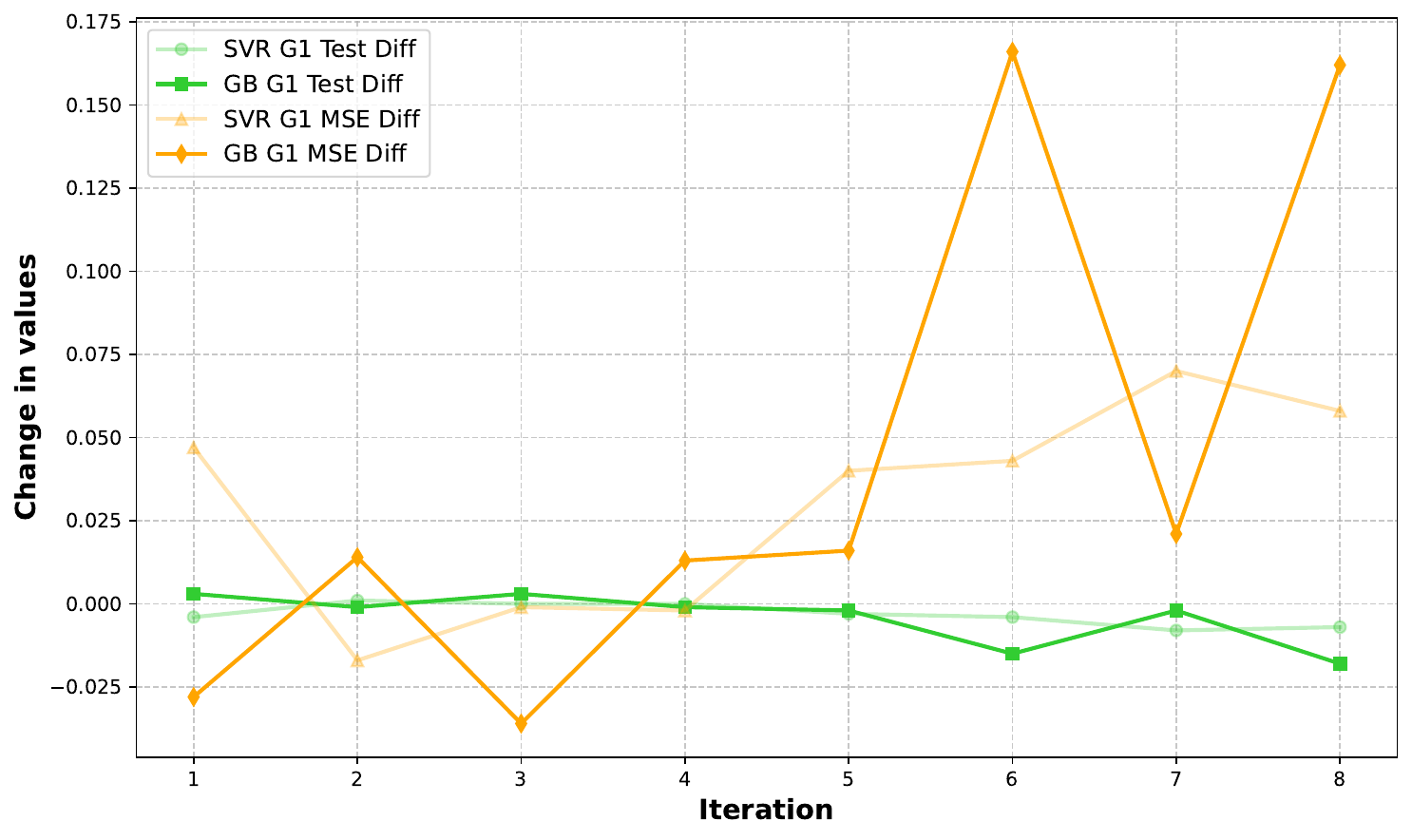}
         \caption{G3}
    \end{subfigure}
    \caption{Reduce\_data\_mutator comparison results in G1 and G3 datasets. Iterations of data mutation is on x-axis, and relative change in values of R\textsuperscript{2} test and MSE on y-axis.}
    \label{fig:reduce_compare}
\end{figure}

For a coherent comparison of the results among data mutators and models, the fluctuation trends in the R\textsuperscript{2} test and the MSE values are visualized in Figures~\ref{fig:reduce_compare}, \ref{fig:select_compare}. Section \ref{sec:effectanalysis} is structured to answer the RQs 1-3 presented in Section \ref{sec:RQs}

\subsection*{RQ1: How does the reduction in training data influence the resilience of the ML model?}
Despite some similarities in the behavior of GB and SVR, in both datasets G1 and G3, GB performed better and showed greater resilience to the reduction in the mutation of the training data. The highest R\textsuperscript{2} test and lowest MSE were attained at the 6\textsuperscript{th} iteration indicating that, for ASOs data, collecting a large training dataset does not necessarily improve the performance of the ML model. In this case, 30\% of the available data were sufficient to achieve the best performance.

In the case of G3 dataset, which is a more noisy and volatile dataset, the GB model showed high resilience in the predicted R\textsuperscript{2} test; however, after the 5\textsuperscript{th} iteration, the MSE fluctuations recorded slightly higher changes. Both models performed closely until the 5\textsuperscript{th}, where 50\% of the training data was removed. In subsequent iterations, the models showed an increase in MSE and a spiking pattern in the case of the SVR model, as shown in \ref{fig:reduce_compare}. The obtained results are expected since the ML model needs more data to be able to sustain good performance in noisy data and generalize without fitting the noise.

\subsection*{RQ2: How does the selection of the size of a certain class of data influence the resilience of the ML model?}
In response to the select\_data\_mutator both models seem to have identical fluctuating behavior in the MSE values. The change in the R\textsuperscript{2} test values is nearly negligible.  As a result, in the G1 dataset, both models show similar unpredictable performance in response to the decrease in one class of the data.

Despite the fluctuating performance in response to the reduction of a specific class from the ASOs data, GB model showed slightly better resilience corresponding to the MSE pattern across iterations in G3 data. The resilience of the models during the first four iterations shows relatively stable behavior, contrary to smaller datasets.

\subsection*{RQ3: How to evaluate the resilience of different ML models in response to different data mutators?}
\begin{figure}
\centering
    \begin{subfigure}{0.4\textwidth}
         \includegraphics[width=\textwidth]{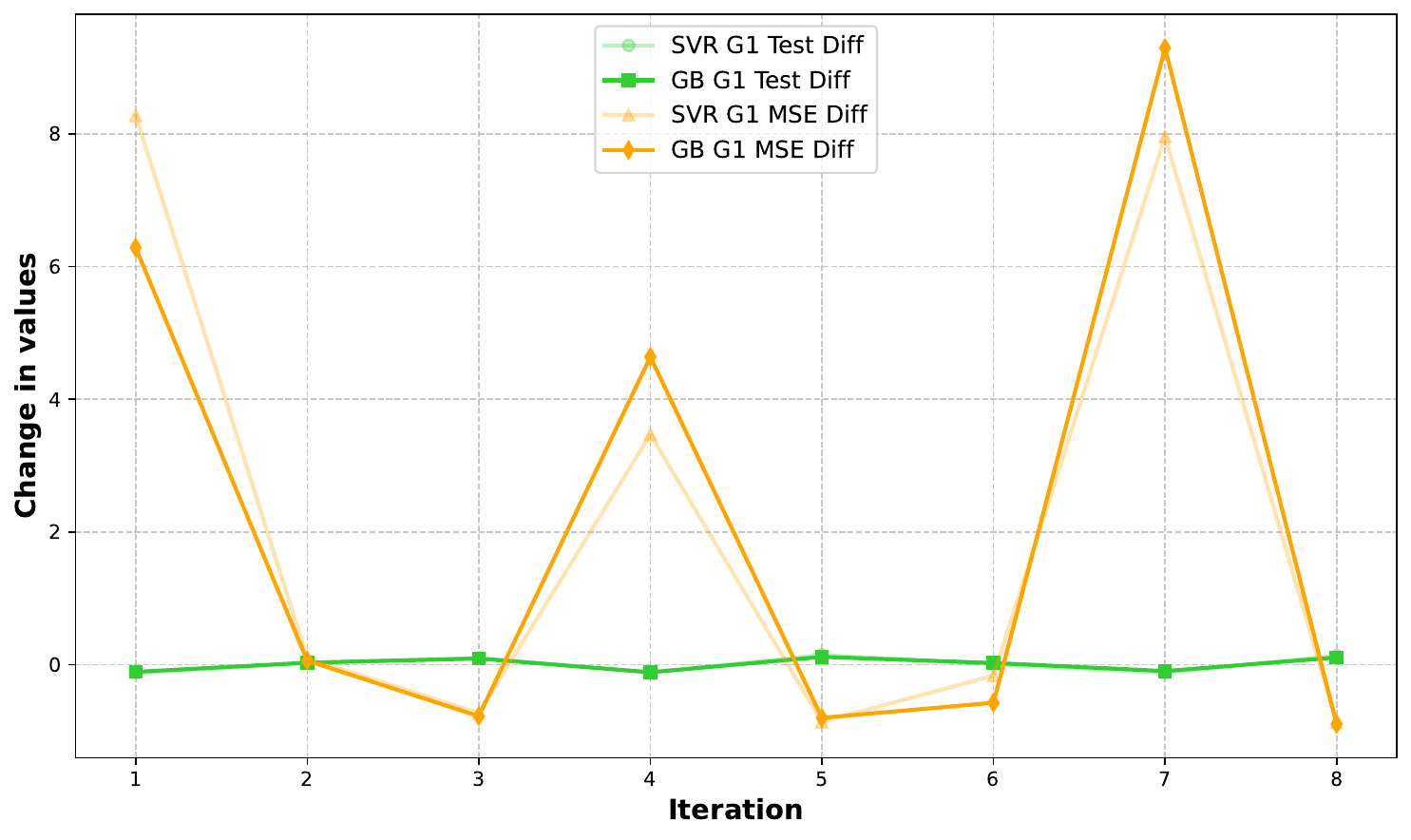}
         \caption{G1}
    \end{subfigure}
    \begin{subfigure}{0.4\textwidth}
         \includegraphics[width=\textwidth]{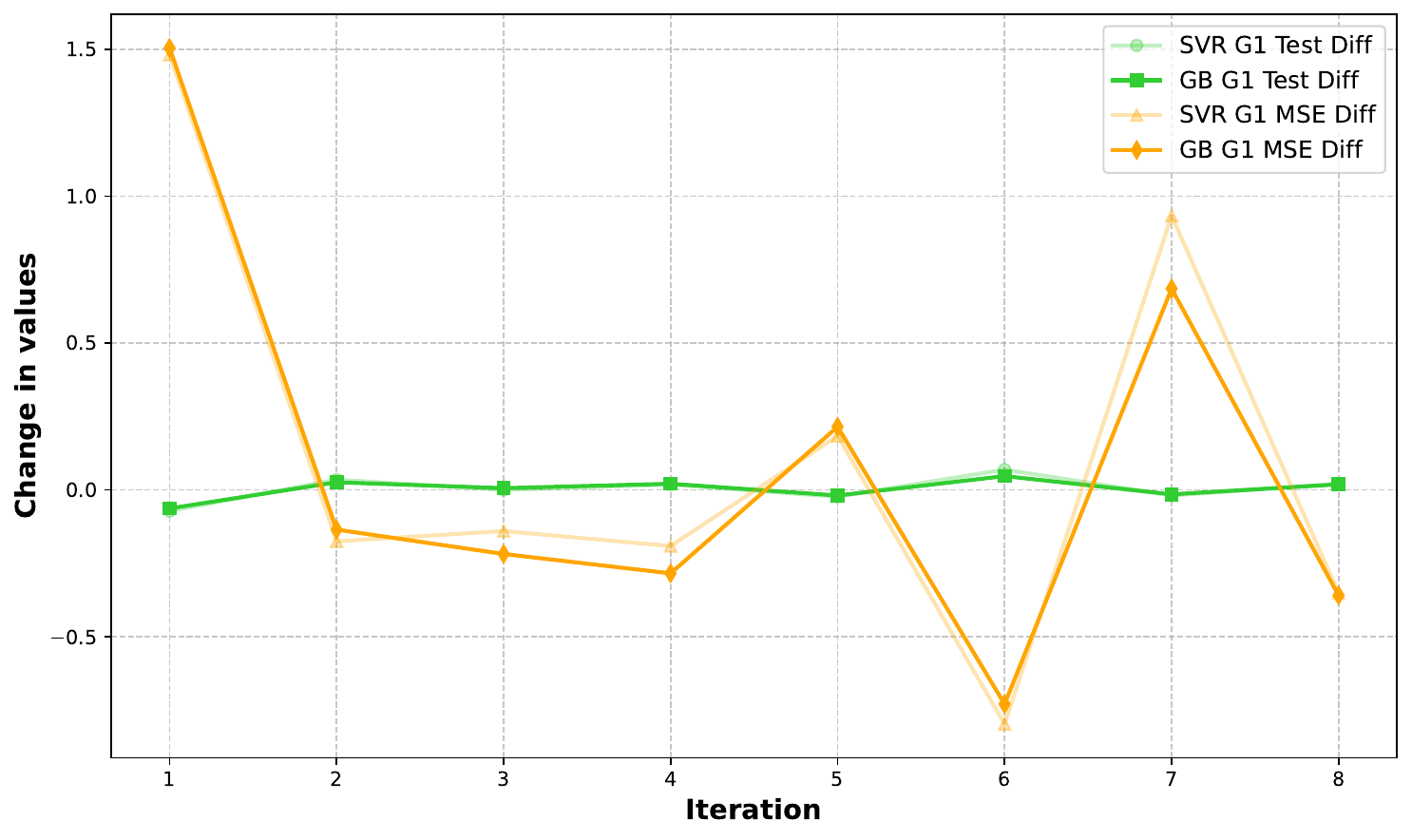}
         \caption{G3}
    \end{subfigure}
    \caption{Select\_data\_mutator comparison results in G1 and G3 datasets. Iterations of data mutation is on x-axis, and relative change in values of R\textsuperscript{2} test and MSE on y-axis.}
    \label{fig:select_compare}
\end{figure}

The fluctuations in MSE values in response to the reduce\_ and select\_data\_ mutators were more significant than that of the R\textsuperscript{2} test as shown in Figure \ref{fig:reduce_compare} and Figure \ref{fig:select_compare}. Therefore, MSE shows higher sensitivity to data mutation changes compared to other investigated metrics.

\section{Conclusion}\label{sec:conclusion}
To test the resilience of ML models to multiple intentionally triggered faults, we present a Fault Injection for Undesirable Learning in the input data (FIUL-Data) testing framework. The proposed framework is evaluated on a case study of ASOs data where the performance of GB and SVR models is compared for each data mutator. In response to reduce\_data\_mutator, both models show relatively high resilience in larger datasets with GB outperforming SVR. We observe that 30\% of the available data were sufficient to achieve the best performance. Regarding the second type of mutation, the models show greater resilience in the G3 dataset with a decreasing trend of MSE compared to G1, which had unpredictable performance. This shows that when the size of the -P=O data is smaller, an ML model performs better. The highest R\textsuperscript{2} test and lowest MSE were achieved at the sixth iteration of the reduced data mutation. Thus in the case of ASOs data, collecting a large training dataset does not necessarily improve the performance of the ML model. Regarding the evaluation metrics, MSE is considered as a sensitive metric since small data mutation significantly changed the MSE behavior.  Therefore, we recommend monitoring the MSE metric when testing the resilience of ML models to data mutations.

For generalization purposes, the FIUL-Data framework could apply to any ML system where the researcher has pre-trained models and can define data mutators. The flexibility of the proposed framework comes from the ability to customize multiple steps depending on the application under study. The FIUL-Data framework can be used in many interesting applications, such as studying how a trained model responds to different kinds of data faults, quantifying and evaluating the trade-off between model resilience and prediction accuracy, and investigating tuning models based on the type of fault in the data.

\section{Threats to Validity}\label{sec:threats}
\subsection*{External Validity}
In any research, generalization of results is important to contribute to the field of study. Despite the application of FIUL-Data on a use case from the analytical chemistry field, the general and flexible design of the proposed framework allows its application in many domains. At every stage of the framework, the user could customize the steps to suit the use case. For example, the data mutators designed in this paper could be modified as the user sees convenient. The ML models applied were trained and optimized for these ASOs datasets; in other applications, other ML models could be studied and compared. 

\subsection*{Reproducibility of Results}
To ensure the reproducibility of the results, we provide a detailed description of the methodology and the experimental setup. The controlled random split of the train and test sets supports reproducibility.

\subsection*{Selection of Datasets}
The evaluation results of the FIUL-Data framework depend on the datasets used. The datasets are generated based on specific and controlled experiments conducted in the Chemistry Department of Karlstad University. Therefore, the ASO compounds are limited to the sequences purchased for the purpose of the experiment. This kind of data has a special characteristic: up to 3 compounds could be derivatives from the same original compound sequence. Therefore, an ASO compound and the derivative compounds resulting from the separation process share similar characteristics, such as phosphorothioation and the frequency of nucleotide bases in the sequence.  Since these characteristics are transformed into features, the underlying similarity could impact the models' performance during the mutation process despite the data's shuffling and random split before the ML application.

\subsubsection{Acknowledgements} 
This work has been funded by the Knowledge Foundation of Sweden (KKS) through the Synergy project - Improved Methods for Process and Quality Controls using Digital Tools (IMPAQCDT) grant number (20210021). In this project, we acknowledge Gergely Szabados, Jakob Häggström, and Patrik Forssén from the Department of Engineering and Chemical Sciences/Chemistry at Karlstad University for their contribution to the acquisition and preprocessing of data.

%
%
%
\bibliographystyle{splncs04}
\bibliography{refs}
%

\end{document}